\documentclass[letterpaper, 10 pt, conference]{ieeeconf}  

\IEEEoverridecommandlockouts                              

\usepackage{graphics} 
\usepackage{epsfig} 
\usepackage{mathptmx} 
\usepackage{times} 
\usepackage{amsmath} 
\usepackage{amssymb}  
\usepackage{amsfonts} 
\usepackage{lipsum}
\usepackage{cite}

\title{\LARGE \bf
Wheelchair Maneuvering with a Single-Spherical-Wheeled Balancing Mobile Manipulator
}

\author{Cunxi Dai$^{1*}$, Xiaohan Liu$^{1*}$, Roberto Shu$^{1}$, and Ralph Hollis$^{1}$
\thanks{*Equal contribution.}
\thanks{$^{1}$The Robotics Institute, Carnegie Mellon University, Pittsburgh, PA 15213, USA
        {\tt\small {cunxid,xiaohan5,rhollis}@cs.cmu.edu, rshumanosalvas@gmail.com.}}%
}

\begin{document}

\maketitle
\thispagestyle{empty}
\pagestyle{empty}

\begin{abstract}

In this work, we present a control framework to effectively maneuver wheelchairs with a dynamically stable mobile manipulator. Wheelchairs are a type of nonholonomic cart system, maneuvering such systems with mobile manipulators~(MM) is challenging mostly due to the following reasons: $1) $ These systems feature nonholonomic constraints and considerably varying inertial parameters that require online identification and adaptation. $2)$ These systems are widely used in human-centered environments, which demand the MM to operate in potentially crowded spaces while ensuring compliance for safe physical human-robot interaction~(pHRI). We propose a control framework that plans whole-body motion based on quasi-static analysis to maneuver heavy nonholonomic carts while maintaining overall compliance. We validated our approach experimentally by maneuvering a wheelchair with a bimanual mobile manipulator, the CMU ballbot. The experiments demonstrate the proposed framework is able to track desired wheelchair velocity with loads varying from 11.8~kg to 79.4~kg at a maximum linear velocity of 0.45~m/s and angular velocity of 0.3~rad/s. 
Furthermore, we verified that the proposed method can generate human-like motion smoothness of the wheelchair while ensuring safe interactions with the environment.


\end{abstract}

\section{INTRODUCTION}

Recent advancements in mobile manipulators have greatly enhanced their utility for assisting humans. A critical aspect in this area is the ability of MMs to operate nonholonomic cart systems, such as shopping carts, luggage carts and hospital beds. These systems usually have one or two caster wheels and a set of fixed wheels that introduce nonholonomic constraints to their motion. They are essential for transporting cargo or human in environments centered around human activities, including supermarkets, hospitals, and construction sites. Among these, pushing a wheelchair to a specified location is vital for individuals with mobility issues, particularly in places like public transport and healthcare centers. Using robots for wheelchair assistance can significantly reduce caregivers' burden and enhance mobility for wheelchair users. When pushing the wheelchair, the robot needs to maneuver the wheelchair with smooth motion while ensuring safe interaction with others around.

Previous works have been done in maneuvering nonholonomic systems with various platforms. 
\cite{nozawaControllingPlanarMotion2012,vazMaterialHandlingHumanoid2020,hawleyControlStrategyImplementation2017} used bipedal humanoid robots to maneuver heavy objects or carts by computing zero momentum points~(ZMP) of interaction with the object and foot placement planning. ~\cite{polveriniMultiContactHeavyObject2020} studied how to change the robot's posture and body leaning angle based on kinematics to maximize force exertion. While being able to exert large force via configuration change, bipedal robots face challenges of the inherent discreteness in their locomotion. This can lead to non-smooth movements of the object being pushed were not carefully controlled, which is not desirable for wheelchair pushing as it may compromise passenger comfort and is less energy-efficient on flat surfaces compared to wheeled robots.

\begin{figure}[t!]
    \centering
    \includegraphics[width=\columnwidth]{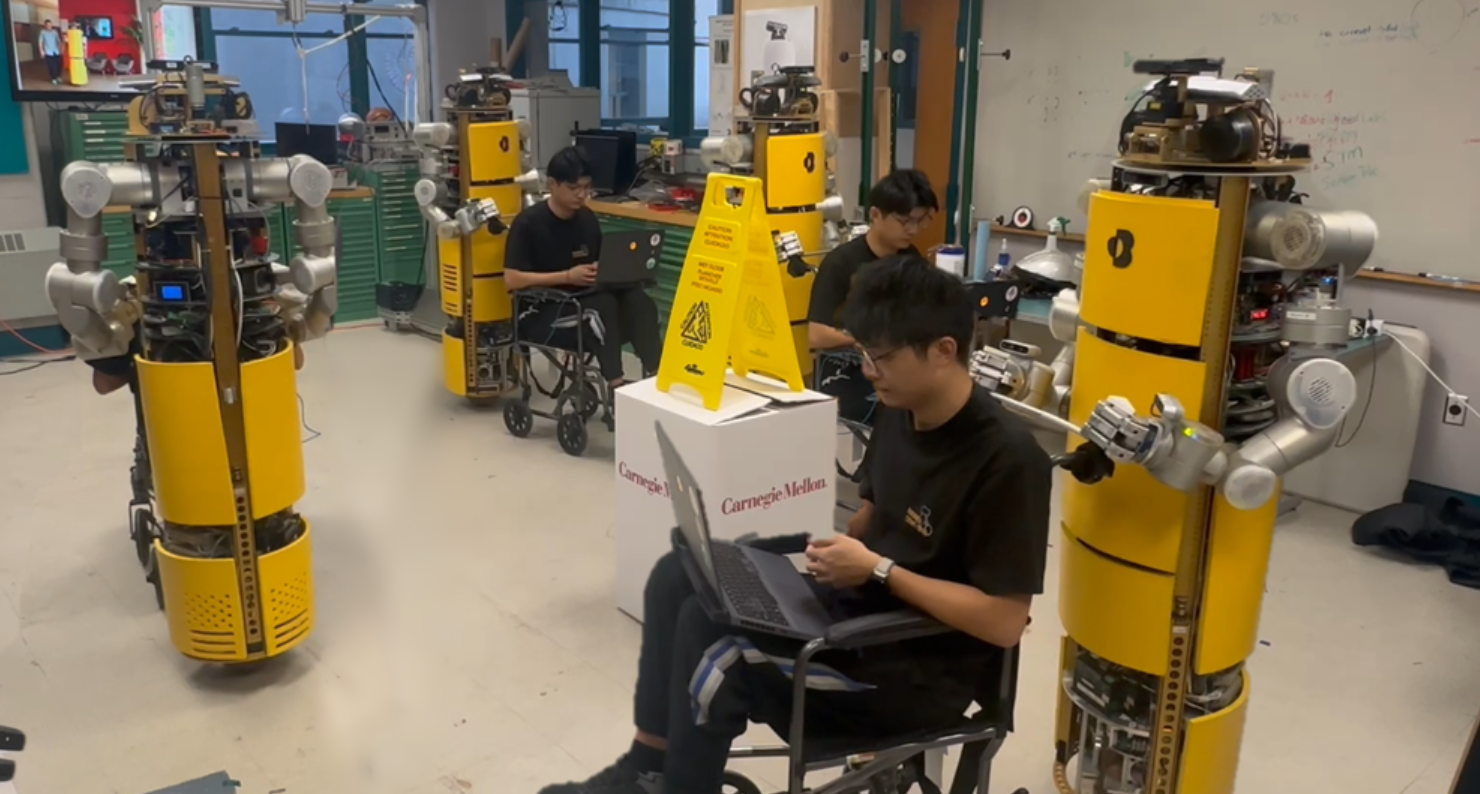}
    \caption{ Time-lapse picture of the CMU ballbot maneuvering wheelchair around obstacle.}
    \label{fig:demo}
    \centering
    \vspace*{-3mm}
\end{figure}

Multi-wheeled mobile base robots are also widely used in cart-pushing tasks.  \cite{5980288,schulzeTrajectoryPlannerMobile2023} studied path planning methods for bimanual mobile-based robots when navigating with a cart, which assumed that the cart can be reoriented by the MM as needed, which does not hold for heavy carts such as wheelchairs with a person. \cite{aguileraControlCartLikeNonholonomic2023, methilPushingSteeringWheelchairs2006, methil2008} proposed trajectory tracking controllers for non-holonomic cart systems that utilize the change of grasping point on the carts to maneuver with a single robot arm. \cite{fujimotoImprovementMethodCompliance2010} proposed an improved impedance controller for cart-pushing tasks to enhance wheelchair stability and ride quality. These results in these studies indicate that multi-wheeled mobile base robots are capable of producing smoother movements compared to bipedal humanoid robots discussed above. However, a human-height wheeled robot can easily become dynamically unstable if it takes a large impact or accelerates too quickly~\cite{1642139}, which led to the adoption of heavy bases with a large footprint in the mentioned studies that hinders its mobility in confined areas like hallways such as hallways.

\begin{figure*}[t!]
    \centering
    \includegraphics[width=2\columnwidth]{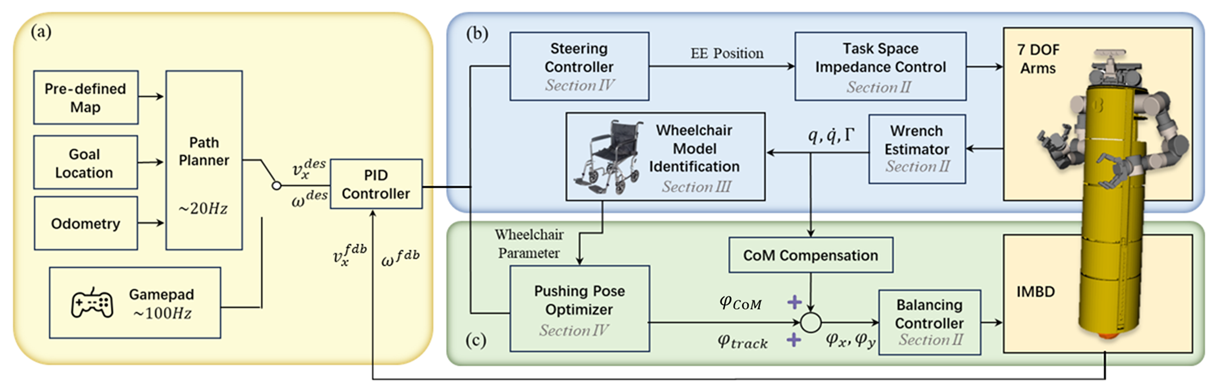}
    \caption{Control framework diagram. (a)~The reference velocity is first generated by the path planner or sent from a gamepad. The velocity is then tracked by the wheelchair-pushing controller by computing the optimal CoM leaning angle and end-effector position commands. (b)~The lower-level task-impedance arm controller then tracks the desired end-effector position commands. A wrench estimator is implemented to provide force estimation that is used for wheelchair parameter identification. (c)~The CoM leaning-angle command is tracked with a balancing controller with CoM compensation. 
 }
    \label{fig:control_diagram}
    \centering
    \vspace*{-3mm}
\end{figure*}

Moreover, previous works lack overall system compliance, which is the compliance in both the arms and the base, either due to hardware limitation or control method, which is key to ensure safe pHRI in crowded environments.

For wheelchair-pushing tasks, we need to further consider the riding experience of the passenger and the safety of the people around. To address this, our goal is to develop a wheelchair-pushing controller for the CMU ballbot that allows accurate and smooth velocity tracking while maintaining overall system compliance.


CMU ballbot is a human-sized bimanual robot that operates while balancing on a single spherical wheel, featuring both compliance and omnidirectionality~\cite{shuMomentumBasedWholeBody2021}. By leaning its body, the ballbot can change its Center of Mass(CoM) leaning angle and exert large forces such as the required force for human sit-to-stand assistance \cite{shominSittostandAssistanceBalancing2015}. The torque-sensing 7-DoF arms provide accurate measurement of the interaction force~\cite{shut_development_2019}. These aspects make the CMU ballot suitable for deployment in caregiving scenarios that require safe pHRI.

In this paper, we present a control framework for the ballbot to dynamically maneuver a wheelchair as shown in Fig.~\ref{fig:demo}, ensuring overall compliance. The controller diagram is demonstrated in Fig.~\ref{fig:control_diagram}.  The framework's key elements include a pushing pose optimizer and a steering controller, which together plan a whole-body motion, enabling the ballbot to track the desired wheelchair velocity. To identify the parameters in the wheelchair's model, an extended Kalman filter~(EKF) is utilized for online system identification~(SI). The main contribution of this paper is a holistic scheme for bimanual dynamic balancing robots to maneuver heavy cart-like systems while ensuring overall compliance.


The remainder of this paper is constructed as follows: Section~\ref{sec:DynModCtrl} introduces the modelling of the CMU. Section~\ref{sec:wheelchair} shows the dynamic model of the wheelchair and online parameter identification. The pushing pose optimizer and steering controller for wheelchair pushing is discussed in Section~\ref{sec:PushCtrl}. Finally, we show that the proposed method can accurately track velocity commands with various loads with experiments described in Section~\ref{sec:Experiment}. Finally, conclusions are summarized in Section~\ref{sec:Conclusion}.

\section{Ballbot Model and Control\label{sec:DynModCtrl}}

 This section introduces the ballbot model and implementation ofbalancing controller and arm impedance controller. 
\begin{figure}[!b]
    \centering
    \includegraphics[width=\columnwidth]{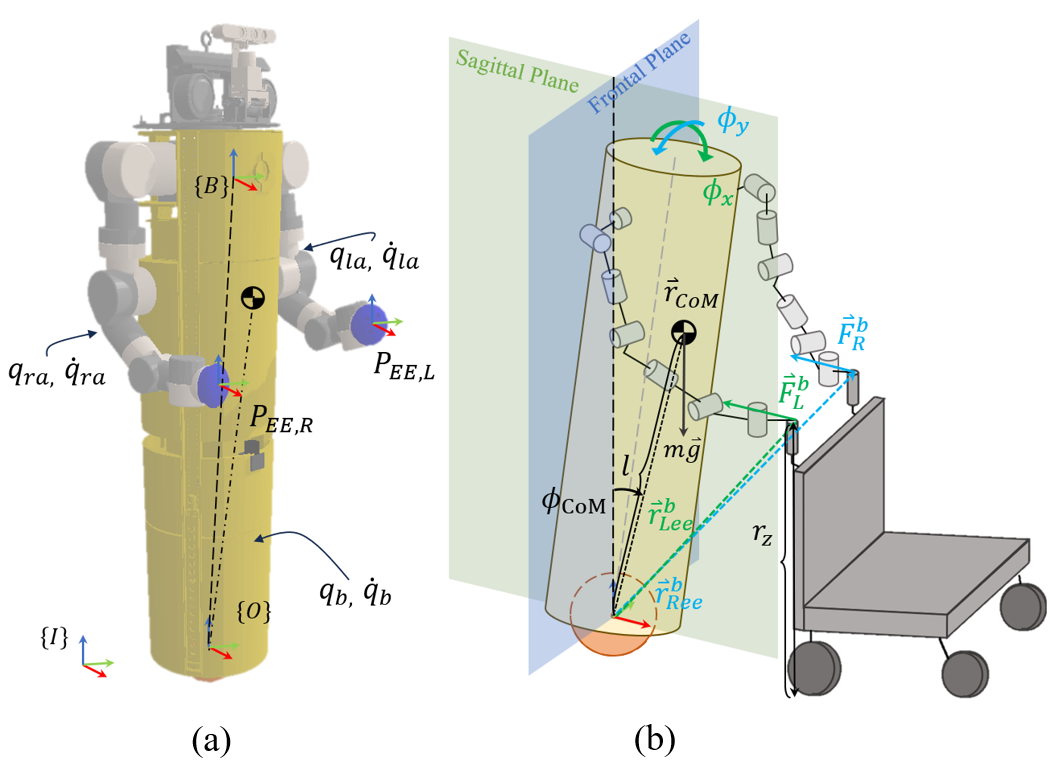}
    \caption{(a)~The ballbot dynamics model. (b)~Quasi-static analysis with forces exerted at the end-effectors.}
    \label{fig:bbdynamics}
    \centering
    \vspace*{-3mm}
\end{figure}

\subsection{Kinematics Model}
The frame choices are shown in Fig.~\ref{fig:bbdynamics}(a), with the origin frame~$\{O\}$ centered on the spherical wheel with axes aligned with the inertial frame $\{I\}$, the body frame $\{B\}$ centered at shoulder center with only its z axis aligned with $\{O\}$. The arm’s task and posture kinematics are defined with respect to body frame $\{B\}$. For both arm, we can write the task relationship between the task coordinates $x\in \mathbb{R} ^{6} $ and the joint configuration coordinates $q\in \mathbb{R} ^{10} $ in the following form:
\begin{equation}
x=\textbf{FK}(q).
\end{equation}
Here, $q=[q_{b},q_{a}]^{T}$. $q_{b} \in \mathbb{R}^{3} $ is the vector of the body pose defined as $[ \phi_{x}, \phi_{y}, \phi_{z}]^{T}$  w.r.t $\{O\}$, where $\phi_{x},\phi_{y}$ are body leaning angles, and $\phi_{z}$ is the body yaw. $q_{a}\in \mathbb{R}^{7} $ is the vector of the arm joints. The task coordinate $x$ is defined as $x = [p^o_x,p^o_y,p^o_z,\phi,\theta,\psi]^T$. Where $[p^o_x,p^o_y,p^o_z]^T$ is the position vector w.r.t $\{O\}$, and $[\phi,\theta,\psi]^T$ are the Z-Y-X Euler angles. The task Jacobian $\boldsymbol{J}(q)$ can then be defined as:
\begin{equation}
\boldsymbol{J}(q)=\frac{\delta \mathbf{F K}(q)}{\delta \boldsymbol{q}} \in \mathbb{R}^{6 \times 10} .
\end{equation}

\subsection{Task-space Impedance Controller}
Another aspect of care-giving tasks is to ensure user's safety while ensuring safe physical human-robot interaction with others. Towards this goal, a task-space impedance controller is used for the arm's controller to track the desired end effector motion while providing stable physical interaction. The desired impedance behavior between external force $F_{e x t}$ and end-effector position error $e_x = x-x_d$ is that of a  mass-spring-damper system of the form:

\begin{equation}\label{MSD}
\bar{M}_d \ddot{e}_x+\bar{B}_d \dot{e}_x+\bar{K}_d e_x=F_{e x t} .
\end{equation}

The symmetric positive definite matrices $\bar{K}_d, \bar{B}_d$, and $\bar{M}_d$ are the desired stiffness, damping, and inertia matrix, respectively. $\boldsymbol{F}_{\text {ext}}$ is the force exerted at the system.
The control law in joint torque space is:
\begin{equation}
\tau^*=\boldsymbol{J}^T\left[\boldsymbol{\Lambda} \ddot{\boldsymbol{x}}_d-\boldsymbol{\Lambda} \overline{\boldsymbol{M}}_d^{-1}\left(\overline{\boldsymbol{B}}_d \dot{\boldsymbol{e}}_x+\overline{\boldsymbol{K}}_d \boldsymbol{e}_x\right)+\boldsymbol{\mu}\right],
\end{equation}
where $\boldsymbol{\Lambda}=\left(J M^{-1} J^T\right)^{-1},~  \boldsymbol{\mu}=\boldsymbol{\Lambda}\left(J M^{-1}\left(\boldsymbol{h}-\tau_{f r i c}\right)-\dot{J} \dot{\boldsymbol{q}}\right)$ where $M \in \mathbb{R}^{10 \times 10}$ is the mass/inertia matrix, $h \in \mathbb{R}^{10}$ is the vector of Coriolis, centripetal, and gravity forces,   $\tau_{f r i c}$ is the joint friction.

\subsection{Balancing Controller}
The balancing controller is a PID controller cascaded with a PD controller that tracks the desired leaning Center of Mass (CoM) leaning angle $\phi_{CoM}$ as shown in Fig.~\ref{fig:bbdynamics}(b). The two arms each have a mass of 12.9 kg, thus posing considerable CoM change to the robot while moving. A CoM compensator is implemented to maintain the CoM to the equilibrium position using body leaning~\cite{shut_development_2019}. 

\section{Wheelchair Modeling \label{sec:wheelchair}}
\subsection{Wheelchair Dynamics}
\begin{figure}[!b]
    \centering
    \includegraphics[width=0.9\columnwidth]{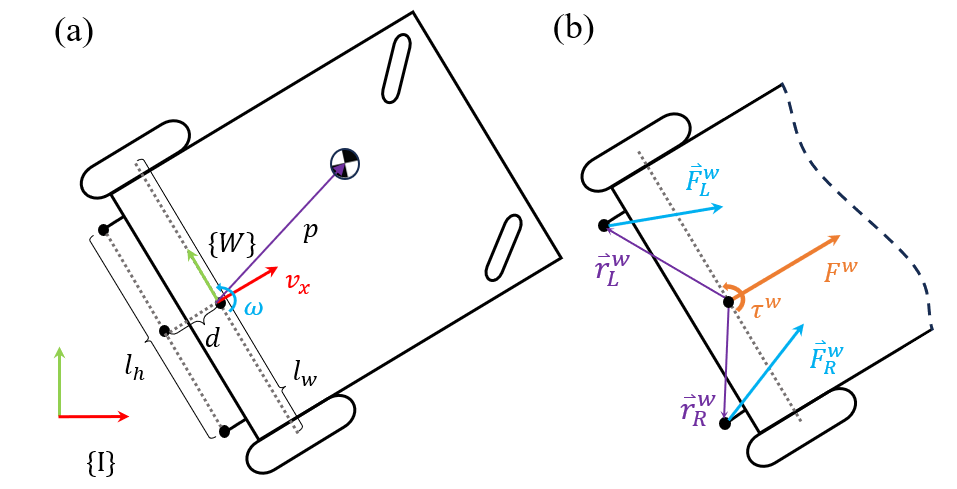}
    \caption{Planar wheelchair model. (a)~Geometric notations. (b)~Input force notations.}
    \label{fig:wcdynamics}
    \centering
    \vspace*{-3mm}
\end{figure}
The wheelchair has two fixed wheels and two caster wheels in the front, since the caster wheels in the front rotate quickly to the direction of motion, we assume that they have a negligible effect on the system's dynamics~\cite{campionWheeledRob17,aguileraModelingInertialParameter2023}. The two rear wheels introduce a nonholonomic constraint such that only motion in the wheelchair's current direction is allowed. A way to model this characteristic is by defining the system velocity as $ \dot{q_w} = \left[v_x,\omega\right]$ w.r.t the wheelchair frame $\left \{ W \right \}$. We can then integrate $\dot{q_w}$ and get the wheelchair's position w.r.t the world frame $\left \{ I\right \}$. Also due to the same constraint, the input wrench exerted at the handle has only 2 effective DoFs, which are the applied force along the x~axis of the cart and torque about the z~axis, as any lateral force will be dissipated by the constraints~\cite{aguileraModelingInertialParameter2023}. Finally, we assume that the wheelchair is always on the floor, thus we can faithfully model the wheelchair's dynamics as a secondary planar system model:
\begin{equation}
\label{projecteddyn}
M_w \ddot{q_w}+C_w(\dot{q_w}) \dot{q_w}=\Gamma_w-N_w(\dot{q_w}),
\end{equation}
where $M_w(q_w)$ is the inertia matrix,  $C_w(\dot{q_w}) $ is the Coriolis term, $N_w( \dot{q_w})$ is the non-conservative force such as the viscous force as the wheels rotate, and $\Gamma_w$ is the external wrench applied by the ballbot. The formulations can be given as: 
$$
\begin{aligned}
& M_w=\left[\begin{array}{ccc}
m_w & 0\\
\frac{m_wp_y(I_w+m\left(p_x^2+p_y^2\right))}{I_w}  & I_w + m_w\left(p_x^2+p_y^2\right) \\

\end{array}\right], \\
& C_w=\left[\begin{array}{ccc}
0 &  -m_w \dot{\boldsymbol{\theta}}p_x \\
0 & 0  \\

\end{array}\right], \quad \Gamma_w=\left[\begin{array}{c}
f_p  \\
\tau
\end{array}\right],  \quad N_w=\sigma\left[\begin{array}{c}
\dot{x} \\
\dot{\theta}
\end{array}\right],
\end{aligned}
$$
where $\sigma$ is a positive viscous coefficient that opposes the system's motion, $m_w$ and $I_w$ are the mass and rotational inertia of the wheelchair respectively. The geometric notations are shown in Fig.~\ref{fig:wcdynamics}~(a), where $l_{w},l_{h}$ is the distance between two rear wheels and the two handles respectively, and $d$ is the x-axis offset between the handles and wheels w.r.t $\left \{ W \right \}$. The input wrenches are typically exerted at the handles as shown in Fig.~\ref{fig:wcdynamics}~(b), the mapping between $\Gamma_{input}$ and $\Gamma_w$ is :
\begin{equation}
\Gamma_{input}=\left[\begin{array}{c}
\operatorname{Proj}_{x}\vec{F_L^w} + \operatorname{Proj}_{x}\vec{F_R^w} \\
\vec{r_L^w}\times \vec{F_L^w}+\vec{r_R^w}\times \vec{F_R^w}
\end{array}\right].
\end{equation}
In this context, we assume that the friction between the wheel and the floor is always static friction solely arising from the bearing friction in the rear wheels. We can further model the friction coefficient as:

\begin{equation}
\sigma =[\begin{matrix}
  \mu N & \mu N l_{w}/2 
\end{matrix}].
\end{equation}
where $\mu$ is the rotational friction coefficient, and $N$ is the force exerted on the wheel. By assuming that the wheelchair's mass $m_w$ is evenly distributed on the four wheels, the force on each wheel is $N=mg/4$, with $g=9.8~m^2/s$.

\subsection{Wheelchair Parameter Estimation}
An EKF is used to estimate the unknown parameters similar to~\cite{aguileraModelingInertialParameter2023}. We need to estimate the mass $m_w$, CoM position of the system $\left[p_{x}\ p_{y} \right]^{T}$ and the friction coefficient $\mu$. We pick a set of parameters that would be linear in the dynamic equations by picking $\phi=$ $\left[\begin{array}{lllll}m_w & m_w p_x & m_w p_y & I+m_w\left(p_x^2+p_y^2\right) & \sigma \end{array}\right]^T$, these can then be used to calculate the desired parameters.
The EKF state vector then consists of:
\begin{equation}
\hat{\mathbf{x}}=\left[\begin{array}{lll}
\hat{q_w}^T & \dot{\hat{q_w}}^T & \hat{\phi}^T
\end{array}\right]^T,
\end{equation}
And the derivative of the state vector is:
\begin{equation}
\dot{\hat{\mathbf{x}}}=\left[\begin{array}{c}
\dot{\hat{q_w}} \\
\ddot{\hat{q_w}} \\
\dot{\hat{\phi}}
\end{array}\right]=\left[\begin{array}{c}
\dot{\hat{q_w}} \\
\hat{M_w}^{-1}\left(\Gamma_w-\hat{C_w} \dot{\hat{q_w}}-\hat{N_w}\right) \\
0
\end{array}\right].
\end{equation}
We used a end-effector wrench estimator developed in previous work~\cite{shu_agile_nodate} to estimate the end-effector wrench $\Gamma$ based on joint torque sensor readings. Then a simple moving average filter is applied to smooth results.
The initial values are set to $\phi_0 = \left[\begin{array}{lllll}60 & 0 & 0 & 30 & 0.001\end{array}\right]^T$, and they are updated online at 100~Hz. The result of online estimation during experiment is shown in Section V.

\section{Wheelchair Pushing Controller\label{sec:PushCtrl}}

In this part, we proposed a pushing pose optimizer based on a quasi-static analysis of the system. The key insight of our controller is to treat the arms as passive spring-like elements and use the body leaning as the dominant factor in velocity control. We also introduce a steering controller that helps find optimal reference positions for end-effectors and minimize the required leaning angle of CoM.


\subsection{Pushing Pose Optimizer}

\begin{figure}[t!]
    \centering
    \includegraphics[width=\columnwidth]{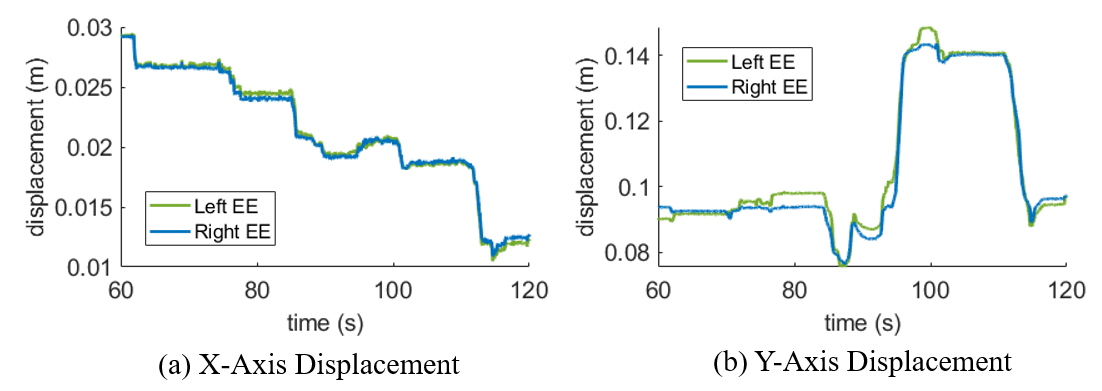}
    \caption{End-effector displacement during 1-minute wheelchair maneuvering experiment on (a)~x axis and (b)~y axis w.r.t $\{B\}$. The maximum displacement mismatch between the two EEs are 0.011m on the x-axis and 0.052m on the y-axis.}
    \label{fig:ee_dis}
    \centering
    \vspace*{-3mm}
\end{figure}

When the ballbot leans, the arms of the robot will be compressed and have force outputs as shown by Eq.~(\ref{MSD}). If the arm controllers share the same parameter, they will have comparable compression thus similar force output. We validated this effect by measuring EE displacement during a 1-minute run as shown in Fig.~\ref{fig:ee_dis}.  To simplify the control, we assume that the two arms have equal output $\vec{F_L^b} = \vec{F_R^b} = \vec{F^b}$. Since we set the rotational stiffness to be 0, we also assume that there is zero output torque. Then the torque equilibrium at the ball center can be expressed as:


\begin{equation}
-\left(\vec{r_{Lee}^b}+\vec{r_{Ree}^b}\right) {\times} F^{b}+ \vec{r}_{\text {CoM }} \times m_{robot} \vec{g} - \tau_{robot} = 0.
\end{equation}

Here, $\vec{r_{Lee}^b}$ and $\vec{r_{Ree}^b}$ are the position vector of end-effectors w.r.t $\left \{ B \right \}$ as shown in Fig.~\ref{fig:bbdynamics}, $m_{robot}$ is the mass of the ballbot, and $\tau_{robot}$ is the net torque result from the friction between the ballbot body and its inverse mouseball drive system (IMBD).

\begin{figure}[!b]
    \centering
    \includegraphics[width=\columnwidth]{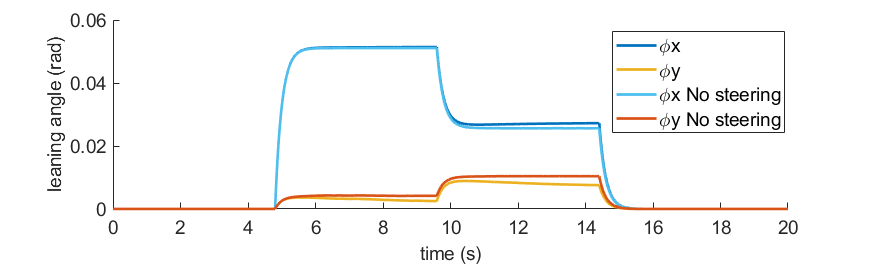}
    \caption{Comparison of ballbot CoM leaning angle $\phi_x,\phi_y$ calculated by the pushing pose optimizer with and without the steering controller. With the steering controller activated, the required CoM leaning angle becomes smaller, indicating less aggressive body movement. }
    \label{fig:compare_steer}
    \centering
    \vspace*{-3mm}
\end{figure}
The EEs are commanded to maintain constant height at the handle, so we can assume that no z-axis force will be exerted. Then, for the x and y axes we have:

\begin{equation}
\begin{aligned}
& 2F^{b}_x r_z=m_{robot} g l \sin \phi_x, 
& 2F^{b}_y r_z=m_{robot} g l \sin \phi_y.
\end{aligned}
\end{equation}

Here, $r_z$ is the height of the wheel chair handle, $F^b_{x}$ and $F^b_{y}$ are the projection of $F^b$ on the sagittal and frontal plane. Linearizing the equations at ${\phi_x,\phi_y}$ = 0, and we can have a mapping between the output force and body leaning angle:
\begin{equation}\label{bodylean}
\begin{aligned}
&F_{x}^{b}= \frac {m_{robot} g l \phi_x}{2 r_z}, \quad 
&F_{y}^{b}= \frac {m_{robot} g l \phi_y}{2 r_z}.\\
\end{aligned}
\end{equation}


Now we move on to the wheelchair system. From Eq.~(\ref{projecteddyn}), we know that when the wheelchair moves at constant velocity $\dot{q_w}$, we will have:

\begin{equation}\label{forceeq}
C_w(\dot{q_w}) \dot{q_w} + N_w( \dot{q_w})=\Gamma_{input},
\end{equation}

Then, expressing the input matrix in ballbot frame we have:

\begin{equation}
\label{forceanglemapping}
\Gamma_{input}=\left[\begin{array}{c}
2\operatorname{Proj}_{x} R^w_b \vec{F^b}\\
(\vec{r^w_L} + \vec{r^w_R}) \times R^w_b \vec{F^b}
\end{array}\right].
\end{equation}

Here, matrix $R^w_b$ is the rotation matrix that transforms a vector in the ballbot frame to the wheelchair frame $\left \{ W \right \}$.


\begin{figure}[t!]
    \centering
    \includegraphics[width=0.6\columnwidth]{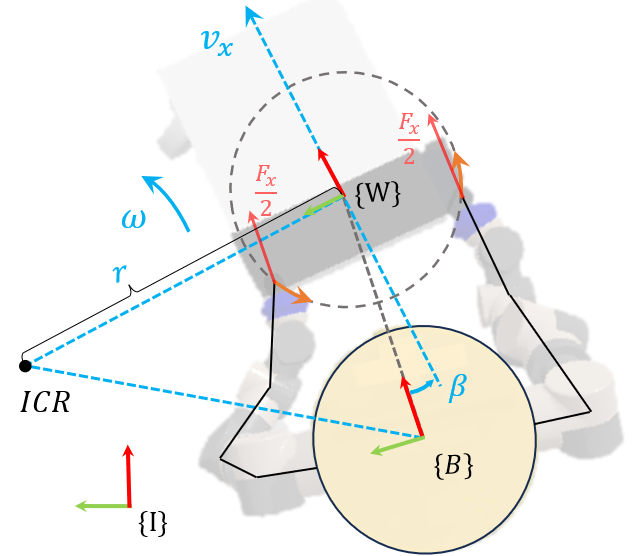}
    \caption{Wheelchair steering controller schematic. The end-effectors steer the wheelchair around the rotational center of wheelchair, i.e. the midpoint of the two rear fixed wheels, based on steering angle $\beta$.}
    \label{fig:steering}
    \centering
    \vspace*{-3mm}
\end{figure}

\begin{figure*}[t!]
    \centering
    \includegraphics[width=2\columnwidth]{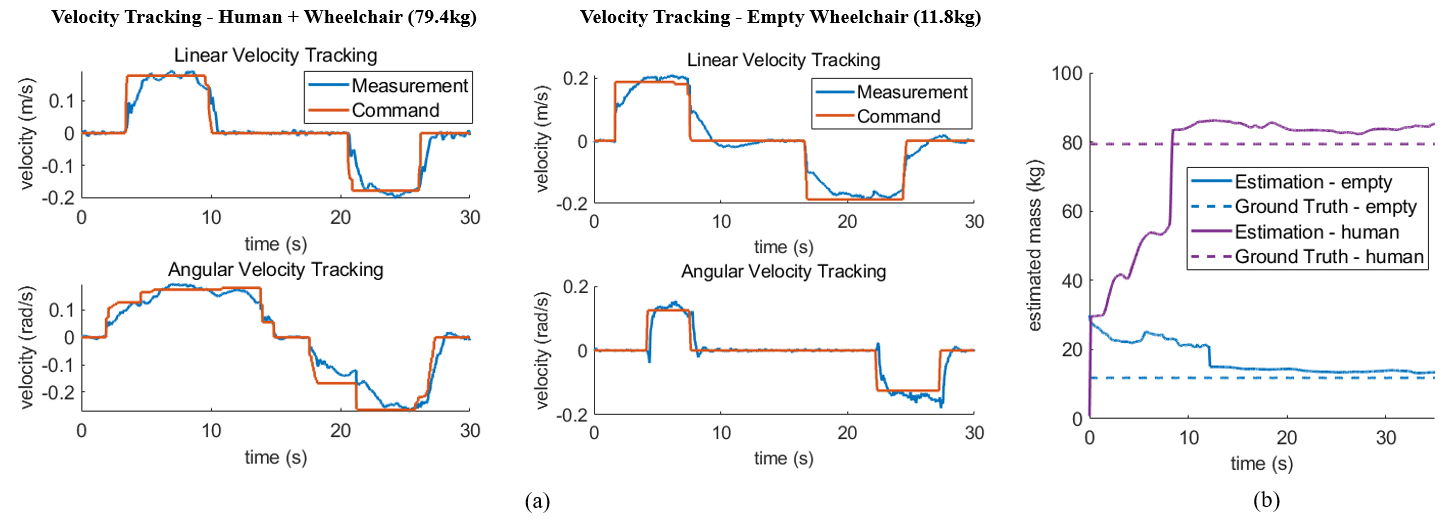}
    \caption{(a)~Velocity tracking performance. The left figure shows results when a human sits in the wheelchair and the right figure shows results with a empty wheelchair. (b)~Online mass estimation with human (79.4~kg in total) and empty cart~(11.8~kg).}
    \label{fig:vel_track}
    \centering
    \vspace*{-3mm}
\end{figure*}

\begin{figure*}[t!]
    \centering
    \includegraphics[width=2\columnwidth]{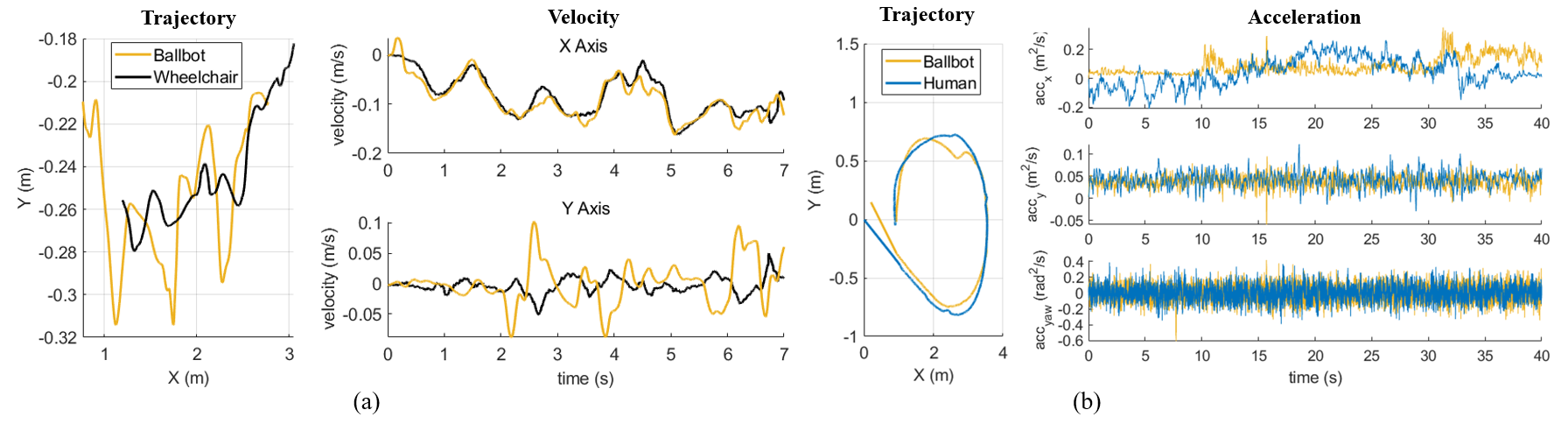}
    \caption{(a)~The trajectory~(left figure) and velocity~(right figure) of both the ballbot and the wheelchair when disturbed while pushing. (b)~The trajectory~(left figure) and acceleration~(right figure) of the wheelchair when pushed with the ballbot and human. }
    \label{fig:disturbance}
    \centering
    \vspace*{-3mm}
\end{figure*}

Based on Eq.~(\ref{forceanglemapping}) , we assume that the output force of the end-effectors is influenced by the leaning angle of the body, $\phi_x$ and $\phi_{y}$. The system is expected to reach the target velocity $\dot{q_w}$ and satisfy the equilibrium equations Eq.~(\ref{forceeq}). Put these equations in matrix form and we have:
\begin{equation}
\underbrace{
  C_w(\dot{q_w}) \dot{q_w} + N_w( \dot{q_w})
}_{\boldsymbol{f}}
=
\underbrace{
    \left[\begin{array}{c}
    2m_{robot}gl\operatorname{Proj}_{x} R^w_b \\
    m_{robot}gl[\vec{r^w_L} + \vec{r^w_R}] _\times R^w_b 
    \end{array}\right]
}_{\mathbf{A}}
\boldsymbol{\Phi},
\end{equation}
where $\Phi = [\phi_{x},\phi_{y}, 0]^T$. $[\vec{a}]_\times$ is the cross-product matrix of $\vec{a}$. For any $\vec{a}, \vec{b}\in \mathbb{R}^3$, we have $[\vec{a}]_\times\vec{b} = \vec{a} \times \vec{b}$.

To calculate the optimal body leaning angle, this problem is formulated as an unconstrained quadratic program problem:
\begin{equation}
\min _{\boldsymbol{f}}(\mathbf{A} \boldsymbol{\Phi} - f)^T Q (\mathbf{A} \boldsymbol{\Phi} - f)
    +\boldsymbol{\Phi}^T \mathrm{R} \boldsymbol{\Phi},
\end{equation}
where $Q, R$ are positive definite weight matrices. The close-form solution is:
\begin{equation}
\widehat{\boldsymbol{\Phi}}=\left(\mathbf{A}^{\top} \mathbf{Q} \mathbf{A}+\mathbf{R}\right)^{-1} \mathbf{A}^{\top} \mathbf{Q} \boldsymbol{f}.
\end{equation}
In practice, this is solved online at 100~Hz. At each time step, the controller will update the optimal lean angles. The lean angles are then tracked by the low-level balancing controller.

\subsection{Steering Controller}

We want to minimize the ballbot lateral leaning angle $\phi_y$ to avoid collision between the arms and the body, while keeping the ability to exert torque on the wheelchair for turning. To address this, we propose a wheelchair steering heuristic that plans desirable EE positions for the pushing pose optimizer such that $\phi_y$ is minimized.

By positioning the ballbot's body such that the center of $\left \{ W \right \}$ is on the x axis of $\left \{ B \right \}$, the z-axis torque exerted on the wheelchair can be controlled with $\phi_x$ and steering angle $\beta$ as shown in Fig.~\ref{fig:steering}. By steering the wheelchair's direction, we effectively change the z-axis torque exerted on the wheelchair such that the system tracks the desired velocity $\left[v_{x}^{des},\omega^{des}\right]$ around the Instantaneous Center of Rotation (ICR). The turning radius can be calculated as $r_{ICP} = v_{x}^{des}/ \omega^{des}$, and steering angle $\beta$ can be calculated as:
\begin{equation}
\begin{aligned}
\beta = \sin^{-1} (\frac{\omega l_w}{4v_x^{des}d}).
\end{aligned}
\end{equation}

The desired end-effector $\widehat{\vec{r_{Lee}^b}}$ and $\vec{r_{Ree}^b}$ positions can be calculated as:
\begin{equation}
\begin{aligned}
\widehat{\vec{r_{Lee}^b}} &= [d, 0]^T + R_{z}(\widehat{\beta})^T l_w/2, \\
\widehat{\vec{r_{Ree}^b}} &= [d, 0]^T + R_{z}(\widehat{\beta})^T l_w/2, \\
\end{aligned}
\end{equation}
where $R_z$ is the rotation around the z axis. From this, the angular velocity can be controlled with steering angle and linear velocity. Due to arm workspace constraint, steering angle is limited as $\beta \in \left [ -35^{\circ} ,35^{\circ}  \right ] $.

\section{Experiments\label{sec:Experiment}}

\subsection{Velocity Tracking with Different Loads}
To demonstrate the velocity tracking performance of the proposed planner, we tested with an empty wheelchair (11.8~kg) and a human subject weighing 67.6~kg (79.4~kg in total). The parameters mentioned in Section III are estimated online and updated to the wheelchair model, and the velocity command is sent from a gamepad. The results are shown in Fig.~\ref{fig:vel_track}, where (a) shows linear and angular velocity tracking with different loads, and (b) shows the online estimated mass. When maneuvering an empty cart, the average system response time is 1.6~s for 0.2~m/s step in linear velocity, and 1.7~s for 0.15~rad/s step in angular velocity. When maneuvering a wheelchair with a person, the average system response time is 1.1~s for 0.2~m/s step in linear velocity, and 1.8~s for 0.18~rad/s step in angular velocity. The online estimation converges in 10.2~s with a 3.2~kg (4$\%$) steady-state error when pushing wheelchair with a human, and converges in 13.4~s with a 1.1~kg (9.3$\%$) when pushing an empty wheelchair. The maximum velocity achieved in our experiments with a load on a wheelchair (34.6~kg in total) is 0.45~m/s of linear velocity and 0.3 rad/s of angular velocity.  Due to the limited space in the room, we did not conduct further tests on the system's maximum speed.  Furthermore,  the system demonstrated the ability to effectively turn the wheelchair in place, which is important to navigating in narrow spaces such as hallways and elevators.

\subsection{Motion Smoothness}
In this experiment, we compared the wheelchair's acceleration when maneuvered by the ballbot against when manuvered by human as a metric to evaluate motion smoothness~\cite{schulze_trajectory_2023}. Initially, we remotely operated the ballbot to navigate through the lab around a obstacle as shown in Fig.~\ref{fig:demo}, followed by a human attempting to replicate the same path. An Intel Realsense T265 sensor is mounted on the wheelchair to record the trajectory and acceleration data as shown in Fig.~\ref{fig:disturbance}~(b). We assessed the wheelchair's motion smoothness using the acceleration norm as a metric, with a lowere acceleration norm suggesting gentler accelerations or decelerations, thereby indicating smoother movements.

The data showed that the ballbot's maneuvering resulted in 25.3\% higher acceleration norm on the x axis, 21.7\% on the y axis, and 19\% on the z axis w.r.t $\{B\}$ compared to human maneuvering. This result showed that our approach can generate human-like motion smoothness of the wheelchair, which is essential for ensuring passenger's comfort.

\begin{figure*}[t!h]
    \centering
    \includegraphics[width=2\columnwidth]{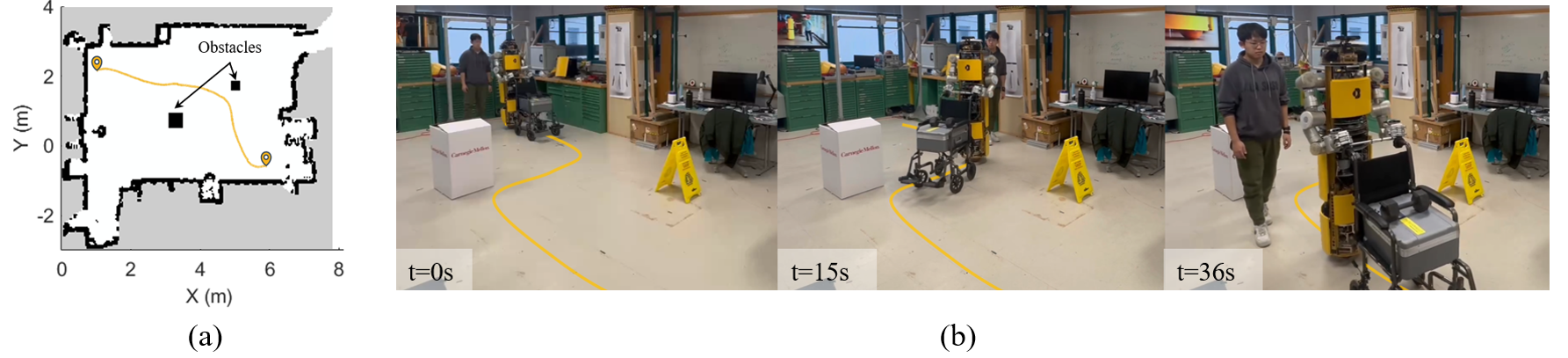}
    \caption{Navigation across lab with predefined obstacles. (a)~Map with pre-defined obstacles. The markers indicates the start position(left marker) and goal position(right marker), and the yellow line shows the actual trajectory of the ballbot. (b)~Time-lapse picture of the experiment.}
    \label{fig:navigation}
    \centering
    \vspace*{-3mm}
\end{figure*}

\subsection{Physical Compliance}
Another important goal of the proposed controller is the ability to ensure safe pHRI with overall compliance. To validate this, we manually disturbed the ballbot's body while it is tracking a linear velocity command of 0.1~$\mathrm{m/s}$. The velocity and position of both the ballbot and wheelchair are measured to analyze how the impact is transmitted to the wheelchair. As suggested in Fig.~\ref{fig:disturbance}~(a), the x-axis velocity of the wheelchair closely mirrors that of the ballbot. On the y axis, the disturbance is more evidently weakened due to overall compliance, resulting in a less aggressive velocity change of the wheelchair compared to the ballbot. Furthermore, we tested the overall compliance of the system by measuring the minimum force required to move the wheelchair by pushing on ballbot. The measured required force is 22.7~N on the x axis and 12.3~N on the y axis w.r.t $\{B\}$ for the system to move, showing that the overall compliance of the system can ensure safety for people that might collide with the system, and can be manually stopped with small amount of force.

\subsection{Navigating with Wheelchair}
A vital task for wheelchair maneuvering is to navigate it to the desired position while avoiding obstacles. In this experiment, we integrated the proposed controller with the ROS navigation framework~\cite{Rosnavigation}. The task is to navigate the wheelchair with a 34.6~kg load across the lab in a known map and avoid predefined obstacles as shown in Fig.~\ref{fig:navigation}(a). The planner replans at 20 Hz frequency online and sends real-time velocity commands to the controller. Acceleration constraints were added to the planner to generate feasible commands for the wheelchair-pushing system and avoid drastic movements. Fig.~\ref{fig:navigation}(b) shows the ballbot's velocity during the experiment, and the time-lapse picture of this experiment is shown in Fig.~\ref{fig:navigation}(c), the system successfully reached the given goal location without collision. This experiment shows the controller's ability to effectively track real-time velocity commands and compatibility with the existing navigation pipeline.

\section{Conclusion\label{sec:Conclusion}}
We present a control framework that enables the CMU ballbot to maneuver a wheelchair while maintain balancing. The core idea is to utilize whole-body motion to maneuver while maintaining overall compliance. The proposed method is evaluated in real hardware experiments, showing that the approach can perform desirable maneuvers while ensuring smooth wheelchair motion and safe pHRI. In future work, autonomous affordance detection will allow grasping point identification. It will also be necessary to evaluate the ballbot pushing a wheelchair up and down ADA-compliant ramps~\cite{7139516}. Furthermore, it is likely that the proposed framework can be applied to similar non-holonomic cart systems such as wheelbarrows and shopping carts.

 \addtolength{\textheight}{-9cm}   




\section*{ACKNOWLEDGMENT}

This work was supported in part by NSF grant CNS-1629757.


\bibliographystyle{IEEEtran}
\bibliography{IEEEabrv,ref.bib}

\end{document}